\documentclass[review]{elsarticle}
\usepackage{lineno,hyperref}
\modulolinenumbers[100]

\journal{}

\usepackage{times}
\usepackage{url}
\usepackage{latexsym}

\usepackage{graphicx}
\usepackage{epstopdf}
\usepackage{url}
\usepackage{multirow}
\usepackage{booktabs}
\usepackage{tikz}
\usepackage{overpic}
\usetikzlibrary{fit,shapes.misc}
\usetikzlibrary{calc,shapes.callouts,shapes.arrows}
\usepackage{tikz}
\usepackage{units}
\usepackage{etex}
\usetikzlibrary{arrows,automata,shapes,snakes,automata,backgrounds,petri,trees,positioning}

\usepackage{array}

\usepackage{multirow}

\usepackage{tkz-tab}
\usepackage{latexsym}
\usepackage{amssymb}
\usepackage{amsmath}
\usepackage{scalefnt}

\usepackage{pgfplots}
\pgfplotsset{compat=newest,compat/show suggested version=false}
\definecolor{halfgray}{gray}{0.55} 
\definecolor{webgreen}{rgb}{0,.5,0}
\definecolor{webbrown}{rgb}{.6,0,0}
\definecolor{blue-violet}{rgb}{0.54, 0.17, 0.89}
\definecolor{darkorange}{rgb}{1.0, 0.55, 0.0}
\definecolor{upmaroon}{rgb}{0.48, 0.07, 0.07}
\definecolor{uscgold}{rgb}{1.0, 0.8, 0.0}

\definecolor{blue(pigment)}{rgb}{0.2, 0.2, 0.6}
\definecolor{capri}{rgb}{0.0, 0.75, 1.0}

\definecolor{cyan(process)}{rgb}{0.0, 0.72, 0.92}
\definecolor{darkblue}{rgb}{0.0, 0.0, 0.55}

\colorlet{Mycolor1}{green!10!orange!90!}
\definecolor{Mycolor2}{HTML}{00F9DE}

\pgfplotsset{
	discard if/.style 2 args={
		x filter/.code={
			\edef\tempa{\thisrow{#1}}
			\edef\tempb{#2}
			\ifx\tempa\tempb
			\def\pgfmathresult{inf}
			\fi
		}
	},
	discard if not/.style 2 args={
		x filter/.code={
			\edef\tempa{\thisrow{#1}}
			\edef\tempb{#2}
			\ifx\tempa\tempb
			\else
			\def\pgfmathresult{inf}
			\fi
		}
	}
}

\usepackage{hyperref}
\usepackage{graphicx}
\usepackage{caption}
\usepackage{subcaption}
\usepackage{tikz}
\usepackage{pgfplots}
\usetikzlibrary{spy,calc}
\usepackage{hyperref}

\newif\ifblackandwhitecycle
\gdef\patternnumber{0}

\pgfkeys{/tikz/.cd,
	zoombox paths/.style={
		draw=orange,
		very thick
	},
	black and white/.is choice,
	black and white/.default=static,
	black and white/static/.style={ 
		draw=white,   
		zoombox paths/.append style={
			draw=white,
			postaction={
				draw=black,
				loosely dashed
			}
		}
	},
	black and white/static/.code={
		\gdef\patternnumber{1}
	},
	black and white/cycle/.code={
		\blackandwhitecycletrue
		\gdef\patternnumber{1}
	},
	black and white pattern/.is choice,
	black and white pattern/0/.style={},
	black and white pattern/1/.style={    
		draw=white,
		postaction={
			draw=black,
			dash pattern=on 2pt off 2pt
		}
	},
	black and white pattern/2/.style={    
		draw=white,
		postaction={
			draw=black,
			dash pattern=on 4pt off 4pt
		}
	},
	black and white pattern/3/.style={    
		draw=white,
		postaction={
			draw=black,
			dash pattern=on 4pt off 4pt on 1pt off 4pt
		}
	},
	black and white pattern/4/.style={    
		draw=white,
		postaction={
			draw=black,
			dash pattern=on 4pt off 2pt on 2 pt off 2pt on 2 pt off 2pt
		}
	},
	zoomboxarray inner gap/.initial=5pt,
	zoomboxarray columns/.initial=3,
	zoomboxarray rows/.initial=2,
	subfigurename/.initial={},
	figurename/.initial={zoombox},
	zoomboxarray/.style={
		execute at begin picture={
			\begin{scope}[
				spy using outlines={%
					zoombox paths,
					width=\imagewidth / \pgfkeysvalueof{/tikz/zoomboxarray columns} - (\pgfkeysvalueof{/tikz/zoomboxarray columns} - 1) / \pgfkeysvalueof{/tikz/zoomboxarray columns} * \pgfkeysvalueof{/tikz/zoomboxarray inner gap} -\pgflinewidth,
					height=\imageheight / \pgfkeysvalueof{/tikz/zoomboxarray rows} - (\pgfkeysvalueof{/tikz/zoomboxarray rows} - 1) / \pgfkeysvalueof{/tikz/zoomboxarray rows} * \pgfkeysvalueof{/tikz/zoomboxarray inner gap}-\pgflinewidth,
					magnification=3,
					every spy on node/.style={
						zoombox paths
					},
					every spy in node/.style={
						zoombox paths
					}
				}
				]
			},
			execute at end picture={
			\end{scope}
			\gdef\patternnumber{0}
		},
		spymargin/.initial=0.5em,
		zoomboxes xshift/.initial=1,
		zoomboxes right/.code=\pgfkeys{/tikz/zoomboxes xshift=1},
		zoomboxes left/.code=\pgfkeys{/tikz/zoomboxes xshift=-1},
		zoomboxes yshift/.initial=0,
		zoomboxes above/.code={
			\pgfkeys{/tikz/zoomboxes yshift=1},
			\pgfkeys{/tikz/zoomboxes xshift=0}
		},
		zoomboxes below/.code={
			\pgfkeys{/tikz/zoomboxes yshift=-1},
			\pgfkeys{/tikz/zoomboxes xshift=0}
		},
		caption margin/.initial=4ex,
	},
	adjust caption spacing/.code={},
	image container/.style={
		inner sep=0pt,
		at=(image.north),
		anchor=north,
		adjust caption spacing
	},
	zoomboxes container/.style={
		inner sep=0pt,
		at=(image.north),
		anchor=north,
		name=zoomboxes container,
		xshift=\pgfkeysvalueof{/tikz/zoomboxes xshift}*(\imagewidth+\pgfkeysvalueof{/tikz/spymargin}),
		yshift=\pgfkeysvalueof{/tikz/zoomboxes yshift}*(\imageheight+\pgfkeysvalueof{/tikz/spymargin}+\pgfkeysvalueof{/tikz/caption margin}),
		adjust caption spacing
	},
	calculate dimensions/.code={
		\pgfpointdiff{\pgfpointanchor{image}{south west} }{\pgfpointanchor{image}{north east} }
		\pgfgetlastxy{\imagewidth}{\imageheight}
		\global\let\imagewidth=\imagewidth
		\global\let\imageheight=\imageheight
		\gdef\columncount{1}
		\gdef\rowcount{1}
		
	},
	image node/.style={
		inner sep=2pt,
		name=image,
		anchor=south west,
		append after command={
			[calculate dimensions]
			node [image container,subfigurename=\pgfkeysvalueof{/tikz/figurename}-image] {\phantomimage}
			node [zoomboxes container,subfigurename=\pgfkeysvalueof{/tikz/figurename}-zoom] {\phantomimage}
		}
	},
	color code/.style={
		zoombox paths/.append style={draw=#1}
	},
	connect zoomboxes/.style={
		spy connection path={\draw[draw=none,zoombox paths] (tikzspyonnode) -- (tikzspyinnode);}
	},
	help grid code/.code={
		\begin{scope}[
			x={(image.south east)},
			y={(image.north west)},
			font=\footnotesize,
			help lines,
			overlay
			]
			\foreach \x in {0,1,...,9} { 
				\draw(\x/10,0) -- (\x/10,1);
				\node [anchor=north] at (\x/10,0) {0.\x};
			}
			\foreach \y in {0,1,...,9} {
				\draw(0,\y/10) -- (1,\y/10);                        \node [anchor=east] at (0,\y/10) {0.\y};
			}
		\end{scope}    
	},
	help grid/.style={
		append after command={
			[help grid code]
		}
	},
}

\newcommand\phantomimage{%
	\phantom{%
		\rule{\imagewidth}{\imageheight}%
	}%
}
\newcommand\zoombox[2][]{
	\begin{scope}[zoombox paths]
		\pgfmathsetmacro\xpos{
			(\columncount-1)*(\imagewidth / \pgfkeysvalueof{/tikz/zoomboxarray columns} + \pgfkeysvalueof{/tikz/zoomboxarray inner gap} / \pgfkeysvalueof{/tikz/zoomboxarray columns} ) + \pgflinewidth
		}
		\pgfmathsetmacro\ypos{
			(\rowcount-1)*( \imageheight / \pgfkeysvalueof{/tikz/zoomboxarray rows} + \pgfkeysvalueof{/tikz/zoomboxarray inner gap} / \pgfkeysvalueof{/tikz/zoomboxarray rows} ) + 0.5*\pgflinewidth
		}
		\edef\dospy{\noexpand\spy [
			#1,
			zoombox paths/.append style={
				black and white pattern=\patternnumber
			},
			every spy on node/.append style={#1},
			x=\imagewidth,
			y=\imageheight
			] on (#2) in node [anchor=north west] at ($(zoomboxes container.north west)+(\xpos pt,-\ypos pt)$);}
		\dospy
		\pgfmathtruncatemacro\pgfmathresult{ifthenelse(\columncount==\pgfkeysvalueof{/tikz/zoomboxarray columns},\rowcount+1,\rowcount)}
		\global\let\rowcount=\pgfmathresult
		\pgfmathtruncatemacro\pgfmathresult{ifthenelse(\columncount==\pgfkeysvalueof{/tikz/zoomboxarray columns},1,\columncount+1)}
		\global\let\columncount=\pgfmathresult
		\ifblackandwhitecycle
		\pgfmathtruncatemacro{\newpatternnumber}{\patternnumber+1}
		\global\edef\patternnumber{\newpatternnumber}
		\fi
	\end{scope}
}

\usepackage{rotating}
\usepackage{tipa}
\usepackage{graphicx}
\usepackage{forest}
\usepackage{tikz-qtree}
\def\ALG{A\textsc{lg}-D\textsc{aridjah}}
\def\HD{H\textsc{adid}}

\definecolor{cyan(process)}{rgb}{0.0, 0.72, 0.92}
\definecolor{darkblue}{rgb}{0.0, 0.0, 0.55}

\begin{document}

\begin{frontmatter}

\title{ Hierarchical Classification for Spoken  Arabic Dialect Identification  using Prosody: Case of Algerian Dialects}

\author[mymainaddress]{Soumia Bougrine}
\ead{sm.bougrine@lagh-univ.dz}

\author[mymainaddress]{Hadda Cherroun}
\ead{hadda\_cherroun@mail.lagh-univ.dz}

\author[mysecondaryaddress]{Djelloul Ziadi }
\ead{djelloul.ziadi@univ-rouen.fr}

\address[mymainaddress]{Laboratoire d'informatique et Math\'{e}matiques - Universit\'{e} Amar Telidji Laghouat, Alg\'{e}rie}
\address[mysecondaryaddress]{Laboratoire LITIS - Universit\'{e} Normandie Rouen, France}

\begin{abstract}

 In daily communications,  Arabs use  local dialects which are hard to identify automatically using conventional classification methods. The dialect identification challenging task becomes more complicated when dealing with an under-resourced dialects belonging to a same county/region. In this  paper,  we start by analyzing statistically Algerian dialects in order to capture their  specificities   related to prosody  information which  are extracted at utterance level after a coarse-grained consonant/vowel segmentation. According to these analysis findings, we propose  a Hierarchical classification approach for spoken Arabic  algerian Dialect IDentification ({\HD}).  It takes advantage  from the fact that dialects have an inherent property of naturally structured into hierarchy. Within {\HD},  a top-down  hierarchical  classification is applied, in which we use  Deep Neural Networks (DNNs) method to build a local classifier for every parent node into the hierarchy dialect structure. Our framework is implemented and evaluated on Algerian  Arabic dialects corpus. Whereas, the hierarchy dialect structure  is deduced from historic and linguistic knowledges.  The results reveal that  within {\HD},  the best classifier is DNNs compared to Support Vector Machine. In addition, compared with a baseline Flat classification system, our  {\HD}  gives an improvement of $63.5$\% in term of precision. Furthermore, overall results evidence the suitability of our prosody-based {\HD} for speaker independent dialect identification  while requiring  less than \textbf{$6$s} test utterances. 
	
\end{abstract}


\end{frontmatter}

\linenumbers

\section{Introduction}

Dialect IDentification (DID)  is the task of recognizing a dialect automatically, it is  part of Natural Language Processing (NLP).  DID is classified into  two main categories: write-based  or spoken-based. Spoken DID systems  are  featured by their complexity compared to the automatic Language IDentification (LID) because it deals with many variations of the same language.


In general, the applications of spoken DID systems  can be broadly divided into  two categories: front-end for human operators and front-end for machines. As front-end for human operators, spoken DID systems  can be useful in routing calls. In fact, to orient the call to human operators who understand the dialect of the caller.  On the other hand, in the category front-end for machines, it is used in many domains  such as: detection/classification of spoken document retrieval,  enhancing  the performance of automatic speech/speaker recognition, or multi-language translation system.



A dialect/language can be distinguished from another by means of   many characteristics extracted  from the speech information levels: acoustic/phonetic, phonotactic, prosodic, lexical  and syntactic~\cite{Li13}.  These  levels are from   the lowest to the highest speech information. Lexical and syntactic   are more discriminative in LID/DID. However, they require a large vocabulary recognizers. Thus, most  LID/DID systems are based on low level features,  acoustic/phonetic and phonotactic, which    perform well when the recording conditions are  controlled and the record quality is good. In contrast, prosodic based systems are less influenced by noise and  channel variations~\cite{rao2015language}. Furthermore, prosodic systems  are more effective for short utterances while the  phonotactic and acoustic features work better for long utterances~\cite{tong2006integrating}.


Most researches on DID  have only been carried out for non-semitic languages. Whereas, little attention has been paid to  spoken Arabic Dialect IDentification (ADID), especially when dialects belong to the same geographical area or country~\cite{Djellab2016} \cite{Palestinian2015}.
Arabic is a semitic language spoken by more than  $420$ million people in $60$ countries  worldwide~\cite{site1}. It has the following variants: Ancient Arabic (AA), Classical Arabic (CA), Modern Standard Arabic (MSA), and Dialectal Arabic (DA)~\cite{EMBARKI08}. As revealed by its name, AA is found in the old literary writings and mainly the poems and it is  no longer used. CA is the language of the Coran, which is the original source  of  grammar and phonetic rules.  MSA is the official language of all Arab countries. It is used in administrations, schools, official radios,  press, some TV programs. DA is often referred to  colloquial Arabic -vernaculars-, which is used in public places, situations of informal  communications, and  social media. There is a large  number 
of Arabic dialects and they are grouped in five categories: Arabian Peninsula, Levantine, Mesopotamian, Egyptian, and Maghrebi~\cite{versteegh97}. Algerian Arabic  dialect is a Maghrebi  dialect. It has  many variations  developed mainly as a result of Arabization phases and  deep colonization history.

In this paper, we propose  a Hierarchical classification approach for spoken Arabic Dialect Identification ({\HD}) where speech is  characterized at prosodic level  using deep learning.
A Hierarchical Classification (HC) is a machine learning method of placing new items into a collection on the light of a predefined hierarchical structure~\cite{silla2011survey}.

 The purpose of our investigation  is three folds. First,  we focus on measuring the discriminative power of the prosody in Algerian Arabic dialects. Indeed,   existing  ADID systems rely   mainly on  knowledges  extracted  from  acoustic/phonetic and  phonotactic cues while dismissing  prosodic ones in spite of their advantages. In fact, it has been proved that dialect variations are notably   pursued in  prosodic features~\cite{wells1982}. 

The second investigation  concerns measuring the effect of    Hierarchical classification in ADID. The main idea behind that is to exploit the fact that the languages, especially  dialects, have an inherent property of naturally structured into hierarchy. Unfortunately, this fact is not taken into  account in the existing works on ADID. Despite the fact that  Arabic dialects are very close and share many linguistics features.  Hence,  their performances decrease quickly when they deal with  more than four dialects.

In the third investigation, we explore Deep Learning   to  build dialect models for ADID system. In fact,  Deep Learning is considered  as state-of-the-art in many  NLP tasks~\cite{collobert2008unified}. It   has  efficient performances for under-resourced speech recognition~\cite{sahraeian2015under}.

The remainder of the paper is organized as follows: in the next section, we review the main existing works on ADID. In Section \ref{Aglance}, we present Algerian dialects, their specificities, and their hierarchical structure derived from some historical  and linguistic studies. In Section \ref{prosoinfo}, we present the prosodic features of speech and  how we extracted them. We also  explain our motivations to leverage prosody for ADID system in this same section. Section \ref{sec:anovaProsodic} is  dedicated to   the statistical analysis of prosody in Algerian Arabic dialects. Afterward, we  describe and explain our  {\HD} approach in Section~\ref{HADIDApp}. The experiments  and results  are described and commented in Section~\ref{experiments}.    Finally, Section~\ref{concl} concludes the paper.

\section{Related Work}
\begin{sidewaysfigure}
	\centering
	\scalebox{1.2}{\input{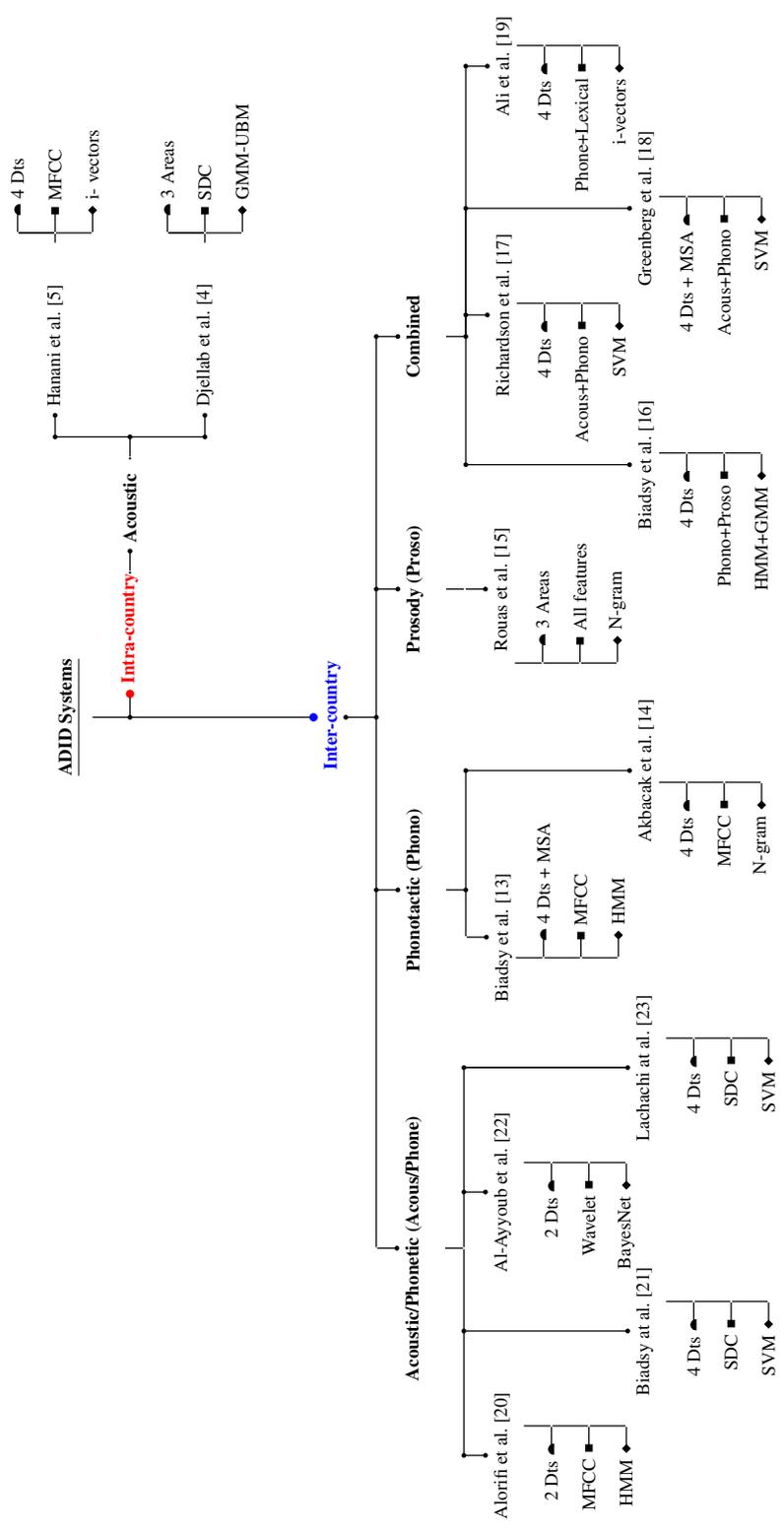}}
	\caption{Taxonomy of ADID systems According to Both Criteria: Speech Feature Level and  Intra/Inter Country.}
	\label{t1}
\end{sidewaysfigure}

We focus in this section on DID systems for Arabic dialects.
The first ADID system was authored by  Rouas et al.~\cite{rouas2006} in $2006$, which is recent investigation  compared to  those  developed for other non-Arabic dialects.
In fact,  for the best of our knowledge,  the pioneer work for non-Arabic dialects appeared  in the middle of the nineties and it is due to  Zissman et al.~\cite{Zissman96}. This lack of interest to ADID is due to many facts. First, there is a noticeable lack of  speech databases/corpora for Arabic dialects dedicated to scientific researches purposes~\cite{bougrine2016toward}~\cite{Zaghouani2014}. Furthermore, there is even less  standard databases ones. For this reason in what follows, we  describe the studied   approaches without considering  their achieved performances intentionally as they deal with different databases.  

In Figure~\ref{t1}, we classify   the main    existing  Arabic DID systems. As a matter of fact, our taxonomy is based on two  criteria: \emph{speech feature Level} and  \emph{Intra/Inter country dialect}. The first criterion indicates from which level  the speech features are extracted. In general, the pre-lexical  levels  are the most used to identify dialect from speech. Hence, we consider acoustic/phonetic, phonotactic and prosodic  levels that  are exploited alone or combined.  The second criterion distinguishes   the origin of targeted dialects in either Intra-country or Inter-country, which means that the studied dialects are  from the same country/region or dialects from many countries. This criterion is chosen because it is more hard to identify Arabic dialects belonging to the same geographical area.

First of all, we summarize  the ADID systems where  acoustic/phonetic models are used alone. Most of the reviewed works are based on spectral features, which are Mel Frequency Cepstral Coefficient (MFCC)
or Shifted Delta Cepstrum (SDC), with Gaussian Mixture Model (GMM)~\cite{Lachachi15} and Support Vector Machine (SVM)\cite{biadsy2011} dialect modeling. 
Alorfi~\cite{Alorfi08} proposed  different acoustic/phonetic approaches  using Hidden Markov Models (HMM). He associated two states for each dialect representing  common and  unique sounds respectively. He  restrained the evaluation of his approach to identify only two Inter-country dialects: Egyptian and  Gulf. 

 Furthermore, Biadsy et al.~\cite{biadsy2011} employed phone labels segmentation to constrain the acoustic models.  They generated dialect  models  using an SVM classifier with special Kernel function, and they  applied this approach on four Arabic Inter-country dialects: Iraqi, Gulf, Levantine and Egyptian.  

In addition to that, Al-Ayyoub et al.~\cite{Al-Ayyoub141} designed an  acoustic model using fixed size segmentation  for which they extracted the selected wavelet features. They deal with two dialects Jordanian  and  Egyptian. 
However, as they confirmed, their results are not conclusive due to the limited size and the quality of the database.
For the context of  Magrebian ADID, Lachachi and Adla~\cite{Lachachi15}  instrumented  the reducing Universal Background Model (UBM)  to support special SVM classification. In fact, they reduced the size of database using the Minimal Enclosing Ball  method by means of a  fuzzy C-mean clustering algorithm. They deal with a database containing  five dialects spoken in: Oran  (Algeria), Algiers  (Algeria), Constantine (Algeria), Morocco and Tunisia. 

In contrast of the other acoustic/phonetic approaches, only Djellab et al.~\cite{Djellab2016} and  Hanani et al.~\cite{Palestinian2015} have proposed ADID system for Intra-country context. 
Hanani et al.~\cite{Palestinian2015} have investigated an acoustic approach based on  i-vectors method for regional accents recognition. They performed their experiments on Arabic Palestinian accents from four different regions: Jerusalem, Hebron, Nablus  and Ramallah. Whereas, Djellab et al.~\cite{Djellab2016}  designed  a GMM-UBM and an i-vectors framework for  accent recognition. They
implement their experiments on a selected data  spoken in  three Algerian ares, which are the East, Center and West of Algeria.




However in phonotactic cues, there are few attempts that have built ADID through the use of a Parallel Phone Recognition followed by Language Modeling (PPRLM)~\cite{Biadsy09Phonotactics} \cite{Akbacak11}. For instance,  
Biadsy et al.~\cite{Biadsy09Phonotactics} have applied  PPRLM using nine (Arabic and non-Arabic) phone recognizers where the Arabic ones are their own built.   
They performed their  experiments on a large database of four Arabic dialects (Egyptian, Gulf, Iraqi, Levantine) together with   MSA.  
More to the point,  Akbacak et al.~\cite{Akbacak11} have designed an approach that combines three models to identify four Arabic dialects (Iraqi, Gulf, Levantine and Egyptian). These models are   cepstral GMM,  PPRLM and   Phone Recognition modeled via SVMs (PRSVM). The combination is carried-out at the score-level. 

Furthermore, some other works have exploited both acoustic/phonetic and phonotactic features to perform  ADID~\cite{NIST12}~\cite{Richardson2009}. Firstly, Greenberg et al.~\cite{NIST12} have designed a combined approach using four core classifiers based on three   spectral similarities and  n-grams. This combination  is done at the back-end level of the system using Bayes classifier. They targeted a set of $24$ languages containing four variations of Arabic language, which are  MSA and three dialects: Iraqi, Levantine and  Maghrebi. Secondly, Richardson et al.~\cite{Richardson2009}  have gathered acoustic and phonotactic features using different classifier. They conduced their experiments on many dialects,  including  Arabic dialects  spoken in  Gulf, Iraq and Levantine. They concluded that SVM classifier has achieved  best results for Arabic dialects.

Ali et al.~\cite{ali2016automatic}  have designed an approach based on  i-vectors method that combined phonetic and lexical features. They performed their  experiments on an Arabic Broadcast speech  database of four Arabic dialects Egyptian, Gulf, Levantine, and North Africa.

On the other side, we have observed that there are few attempts of prosody-based ADID.
%
Based on a previous work of Ghazali et al.~\cite{Ghazali02},  it was shown  that some Arabic dialects (Syria, Jordan, Morocco, Algeria, Tunisia and Egypt) can be  grouped using rhythmic information in three dialectal areas: Maghreb (Morocco, Algeria), Middle-East (Syria, Jordan), and an intermediate one (Tunisia, Egypt). Depending on what was mentioned before,  Rouas et al.~\cite{rouas2006} have designed an ADID system  for three  previous dialectal areas.   Their approach collected all prosodic information:   intonation, rhythm and stress.  Thus, they used a segmentation which is based on consonant/vowel location to get the approximative structure of the syllable. They have  utilized a special codification to represent the duration of phonemes and the energy instead of the real values. Then, they classified dialect areas  using multi-gram models where grams are their pseudo-syllables and each area's dialect is represented by the most frequent sequences of n-gram. Unfortunately, they tested their system  on a small database. 

Another work on ADID has been proposed by  Biadsy et al.~\cite{BiadsyH09prosody}, which  combined the prosodic and phonotactic approaches. In fact, they augmented their  phonotactic system, described above,
by adding some prosodic features like durations and fundamental frequency measured at n-gram level where grams are syllables. They tested their system on four Arabic dialects: Gulf, Iraqi, Levantine, and Egyptian.



To our knowledge there is no deployment of HC in ADID. However,  HLID have been already proposed for others languages. For Indian languages, Jothilakshmi et al.~\cite{jothilakshmi2012hierarchical} designed a two level classification system using acoustic features. On the other hand,  Yin et al.~\cite{yin2007hierarchical}, proposed a   HLID  framework where speech signal is characterized at acoustic level and some prosodic features.   The fusion of these classifiers is performed using  modern GMM fusion system.  Likewise,  Wang et al.~\cite{wang2009hierarchical} suggested a hierarchical system using    bayesian logistic regression models as score  generators. The final identification is performed by  a score based-likelihood  merger.  For Philippine languages, Laguna at al.~\cite{laguna2015development} developed a HLID system via  GMM, in which speeches are characterized by means of   acoustic and prosodic features.

 On the light of this near exhaustive review of the most important ADID systems, 
 let us underline that a little attention has been paid to  Algerian ADID problem. In addition,  we confirm that there is a lack of ADID system that exploit the prosodic information. In fact, only  
 Rouas et al.~\cite{rouas2006} and  Biadsy et al.~\cite{BiadsyH09prosody} have considered this kind of information. 
 Rouas et al.~\cite{rouas2006} have considered area dialects, while Biadsy et al.~\cite{BiadsyH09prosody} have treated inter-country dialects. 
 
 All of the proposed ADID systems use conventional classification method in occurrence SVM, HMM-GMM, BayesNet. However, Deep Learning  is not investigated despite their provided efficiency for Language/Dialect identification \cite{LopezMoreno201646} \cite{Ferrer2016}.

\section{A Glance at Algerian Arabic Dialects}
\label{Aglance}


Algeria is a large country, with a total area of  about $2.4$ million km$^2$. Administratevely divided into $48$ departments, Algeria is bordered  by  mainly three  Arabic  countries, in the north-east by Tunisia, in the east by Libya, in the west by Morocco.  Algeria's official  language  is MSA as in all Arab countries. However,  Algerian local dialects are mostly used instead of MSA. Algerian Arabic is used to refer to dialect spoken in Algeria, known as Daridjah to its speakers.  Algerian dialect presents  a complex linguistic features  mainly due to both Arabization processes that led to the appropriation of the Arabic language by populations  Berber origin, and  the deep colonization. In fact, Arabic Algerian dialect is affected by other languages such as Turkish, French, Italian, and Spanish~\cite{LECLERC2012}.




  According to the Arabization process, dialectologists show that Algerian Arabic  dialects can be divided into two major groups:  Pre-Hil\={a}l\={\i} and Bedouin dialect. Both dialects are different  by  many linguistic features~\cite{marçais1960Algeria} ~\cite{caubet2000questionnaire}.

Firstly,  Pre-Hil\={a}l\={\i} dialect is called  sedentary dialect. It is spoken in areas that are affected  by the expansion of Islam in the $7^{th}$ century.  At this time, the partially affected cities are: Tlemcen, Constantine, and their rural surroundings. The other cities have  preserved their mother tongue language (Berber). Mar{\c{c}}ais \cite{marçais1977esquisse} has divided Pre-Hil\={a}l\={\i} dialect   into two dialects:  village (mountain), and urban dialect.  

\begin{itemize}
	\item   Village dialect is located  between  Trara mountains and Mediterranean  sea. The central town  of this area  is Nedroma, which is located in the  northwestern corner of Algeria.  There is also a village dialect  located in the northeastern corner of Algeria: it is  between  Collo, Djidjelli, and  Mila. 
	
	\item 	Urban dialect is located  in the  northern cites: Tlemcen, Cherchell, Dellys, Djidjelli, and Collo.
\end{itemize}

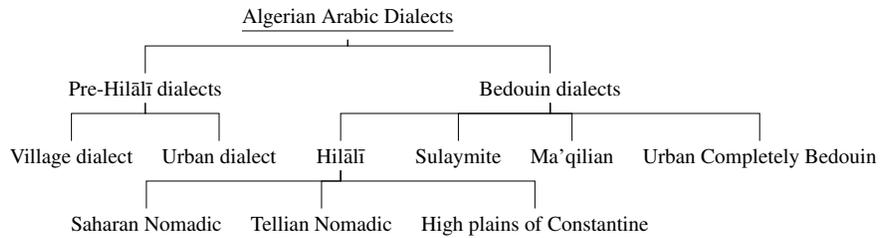
\begin{figure*}[t]
	\centering
	\scalebox{0.8} {\begin{forest}
		for tree={%
			edge path={\noexpand\path[\forestoption{edge}] (\forestOve{\forestove{@parent}}{name}.parent anchor) -- +(0,-12pt)-| (\forestove{name}.child anchor)\forestoption{edge label};}}
		[\underline{{Algerian Arabic Dialects}}
		[{Pre-Hil\={a}l\={\i} dialects}
		[{Village dialect
		}
		]
		[{Urban dialect}
		] ]
		[{Bedouin dialects }
		[{Hil\={a}l\={\i}} [{Saharan Nomadic}
		]
		[{Tellian Nomadic}
		]
		[{High plains of Constantine}
		]
		] 
		[{Sulaymite}
		]
		[{Ma'qilian}
		]
		[{Urban Completely Bedouin}]
		] 
		]
	\end{forest}}
	\caption{Hierarchy Structure for Algerian Dialects.}
	\label{class1}
\end{figure*}

Secondly, Bedouin dialect is spoken in areas which are  influenced by the Arab immigration in the $11^{th}$ century~\cite{PalvaH1},~\cite{PEREIRA1}.  Mar{\c{c}}ais \cite{marçais1960Algeria} has divided  Bedouin dialect into five  distinct basic dialects:
 
 \begin{enumerate}
 	\item Bedouin dialect of eastern Constantine, which are  located in  the region of El Kala and Souf.   It is called 'Sulaymite' dialect because it is connected with Tunisian Bedouin dialects.  
 	
 	\item Bedouin dialect of central and western side of Oran. It is called Ma'qilian dialect because it is connected with Moroccan Bedouin dialects. It covers a part of the arrondissement of Tlemcen, Oran, Sidi Bel Abb\`{e}s, and Sa\"{i}da.

 	\item  Bedouin dialect of the Algerian central and of Sahara. It is  called Saharan Nomadic, it covers almost  the totality of the sahara of Algeria,  towards the east to Oued Righ,  towards the south to the Tadema\"{i}t plateau, and until the west (its limit has not  been clarified).

 	\item Bedouin dialect of the Tell and of the Algerian-Oran Sahel. It is called Tellian Nomadic, that occupies a large part of the Tell of Algeria: Bordj Bou Arr\'{e}ridj, S\'{e}tif, El-Eulma, and El Kantara. 
 
 \item  The dialect of the high plains of Constantine, which covers  the north of Hodna region, and extends to the rough area from Bordj Bou Arr\'{e}ridj to Seybouse river.

 \end{enumerate}

  Mar{\c{c}}ais \cite{marçais1960Algeria} gathered   the  three dialects (3, 4, 5) under the name of  Hil\={a}l\={\i} dialect, which  takes its  name from  Ban\={u} Hil\={a}l tribe. In addition to that, there is  another dialect which has  urban dialects that have been completely influenced   by Bedouin dialect: Annaba, Algiers, B\'{e}ja\"{i}a, Blida,  Mascara, Mazouna, Mostaganem, Medea,   Mila,  Miliana,  Skikda, Tenes, and Oran. For that, we have classified these dialects, so-called  Urban Completely Bedouin (UCB), into the Bedouin dialects group.

  In spite of the sparse and the specific  linguistic  studies that have  dealt with  Arabic dialects, there is no efficient and complete  dialect hierarchy dedicated to Algerian dialects. We have  compiled a preliminary version of such  hierarchy from the above historical knowledges.  We summarize the Hierarchy structure for Algerian dialects  in Figure~\ref{class1}. This Hierarchy structure is also confirmed by some linguistic studies essentially   phonological, lexical and  morphological \cite{marçais1960Algeria} \cite{PalvaH12Classfi06}.  However, this Hierarchy structure has not benefited of deep prosodic analysis for that reason, in what follows, we study which prosodic features are discriminative for Algerian Arabic dialects. 

\section{Prosodic Information for Dialects }
\label{prosoinfo}
In order to identify a language/dialect, many features and measurements are developed in literature to capture prosody inherent to a  speech.   In this section, we first outline our motivation behind the use of prosody for ADID. Then, we describe some  prosodic features related to DID purpose. Finally, we explain how we have extracted them.

 \subsection{Why Prosody for Arabic Dialect Identification? }
 \label{WhyPro}
 
 The knowledge today which we have about  the discriminative power of the  prosody  in Arabic dialects and specially in Algerian ones, can be summarized in what follows:

 \begin{itemize}
 	\item \textit{Arabic dialects differ in their prosodic structure: }  Barkat et al.~\cite{BarkatOP99} evaluated  the discriminating power of prosodic pattern in Arabic dialects in a linguistic study.  
 	They have shown that the prosodic information  can be sufficient to identify Western and Eastern Arabic dialects.

 	\item \textit{Arabic dialects present significant differences at the syllable structure:} Hamdi et al.~\cite{hamdi07} shown that the different types of syllabic structure observed in Arabic dialects can be used as discriminative element.  They demonstrated  that rhythm variation of Arabic dialects is correlated with   syllable structure.  Moreover, Bouziri et al.~\cite{Bouziri1991} confirmed this fact by studying the stress measured through the syllable  structure.

 	\item \textit{Intonation represents a salient  discriminative feature in   Arabic  dialects: } Ghazali et al.~\cite{Ghazali05} studied the nature of intonation of five Arabic dialects. They observed  that the intonation patterns are different   between  Eastern and  Western Arabic dialects. Furthermore, Yeou et al.~\cite{Embarki07} confirmed this result  using another sample of  Eastern and  Western Arabic dialects.

 	\item \textit{Rhythm and intonation are discriminant parameters for some Algerian dialects:}  Benali~\cite{benali2004} studied the role of the rhythm  and  the intonation  in a human identification by means of two Algerian dialects, which are  spoken in Algiers and Oran. He noted that rhythm  and intonation  are very discriminant parameters, particularly the speech rate and the variation of fundamental frequency. 
 \end{itemize}

 On the light of these information, we focus on studying  the rhythm and intonation features  because their importance  to discriminant the Arabic dialects. 

\subsection{Prosodic Features}
\label{prosofeature}


%

To capture prosodic information, the pitch and   duration sequence are used for indicating intonation and rhythm respectively. 


 Rhythm refers to aspects of   temporal organization of speech. To capture quantitative rhythmic variation,  different rhythm metrics have been developed to measure the vocalic and consonantal intervals  in continuous speech. The first and  most  popular metrics used to classify a language include:  Interval Measures (IM)~\cite{Ramus99}, their normalized version (\emph{VarcoV/VarcoC})~\cite{Dellwo06}, and  Pairwise Variability Index (PVI)~\cite{grabe2002}. In addition, we consider  another metric:  \emph{Speech Rate}, which measure the number of syllable per second.

  The IM metrics include  three separate measures:  the duration proportion, the standard deviation of vocalic interval $($\emph{\%V}, \emph{$\Delta$V}$)$,  and the  standard deviation of consonantal interval \emph{$\Delta$C}~\cite{Ramus99}. 
 
 The PVI metrics focus on the temporal succession between the consonantal and  vocalic intervals of the global utterance~\cite{grabe2002}. The model suggests to
  use the raw PVI  for the Consonantal intervals (\emph{rPVI-C}) and  the normalized PVI  for  Vocalic intervals (\emph{nPVI-V}).   	
 


 	

 \begin{table}[!t] 
 	\centering
 	\footnotesize 
 	
 	\begin{tabular}{ p{3cm} p{2cm} p{5.5cm}  }
	
 		\toprule 
 		 Metric   &  Include	  & 	 Definition  \\ 
 		\midrule 
 	
 			 Global Information  & 	\emph{ Bottom }&  	$2^{nd}$ quantiles  pitch nucleus  \\ 
 		& \emph{Top} & $98^{th}$   quantiles  pitch nucleus   \\ 
 		& \emph{Median} & $50^{th}$  quantiles  pitch nucleus   \\ 
 		& \emph{Range} & Difference between \emph{Top} and \emph{Bottom}   \\
 		   \midrule
 		Total Size of Pitch  & \emph{TrajIntra} &  Pitch trajectory of nuclei / duration   \\
 		Trajectory & \emph{TrajInter} & Pitch trajectory between nuclei / duration \\ 
 		\midrule 
 	
	Interval Measures & \emph{\%V}  & The  proportion  of Vocalic interval\\
 		& \emph{$\Delta$V} &  The standard deviation of Vocalic interval    \\
 		& \emph{$\Delta$C} &  The standard deviation of Consonantal interval  \\
 		\midrule
 		Normalized IM & \emph{VarcoV}  & \emph{$\Delta$V} / mean of Vocalic interval duration\\
 	 & \emph{VarcoC}  &  \emph{$\Delta$C} / mean of Consonantal interval duration \\
 		\midrule
 	   Pairwise Variability Index& \emph{rPVI-C} & Raw PVI of  Consonantal intervals  \\  
 	    & \emph{nPVI-V} & Normalized PVI of Vocalic intervals \\  
       \midrule
 	  \emph{Speech Rate} 	& &  Number of syllable per second  \\  
 		\midrule 
%
 		
 	\end{tabular}
 	\caption{  Used Metrics for  Rhythm and Intonation.}
 	\label{table_metric}
 \end{table} 
 The pitch, or fundamental frequency $(F_{0})$, is used for indicating the intonation.  The statistical modeling of intonation are used for each utterance.  We  take into consideration two groups of  intonation metrics:  Global information and total size of pitch trajectory.   Global information 
 metrics  of the pitch are measured  using quantiles of nucleus. This group has four pitch values which are: the \emph{Bottom} is	$2^{nd}$ quantiles  pitch nucleus, the \emph{Median} is $50^{th}$  quantiles  pitch nucleus, the \emph{Top} is  $98^{th}$   quantiles  pitch nucleus in Hz, and  \emph{Range} is the difference between \emph{Top} and \emph{Bottom} in SemiTones (ST). Concerning the total size pitch trajectory, this group has two metrics (TrajIntra, TrajInter) calculated using  pitch trajectory, which is the sum of absolute intervals within/between (TrajIntra/TrajInter)  syllabic nuclei  divided by duration (in ST/s).  Table~\ref{table_metric}  details these features.

 
 \subsection{Segmentation and Features Extraction}
 \label{coarse-grd}
 The extraction of the prosodic features needs a consonant/vowel segmentation. 
 However, for Arabic dialect the problem of phoneme   segmentation is not completely solved and unfortunately there is no related decoders  such as the case for other dialects. To cope with this problem, we rely on the specific Arabic syllable structure to perform a consonant/vowel segment as we explain in what follows:

A syllable is  the unit of pronunciation  between a phoneme and a word. It is  divided into  three components: the opening \emph{Onset}, the central \emph{Nucleus} and the closing  segment  \emph{Coda}. The nucleus and coda are  called the rhyme (or rime)~\cite{crystal08}.	 In Arabic dialect, each syllable   contains at least a   nucleus   while  coda, and onset are optional. The nucleus is  imperatively one or  many vowels,  and the others are consonants.  An Arabic  syllable structure can be modeled by the regular expression  $C^{*}V^{+}C^{*}$  where $C$ is a consonant and $V$ is a vowel.    We exploit this fact to perform a consonant/vowel segmentation. Thus, we  sketched up these measures by considering nuclei as vowel segments and what remains as consonant segments which we call coarse-grained segmentation. For example, we consider the following sequence
 of the three syllables (a). The syllable structure segmentation is presented in (b). Then, we get our coarse-grained segmentation (c), where \emph{VS} (resp. \emph{CS}) represents vowel (resp. consonant) segment.

 \begin{figure}[!h]
 	\centering
 	\includegraphics[width=10cm, height=3cm]{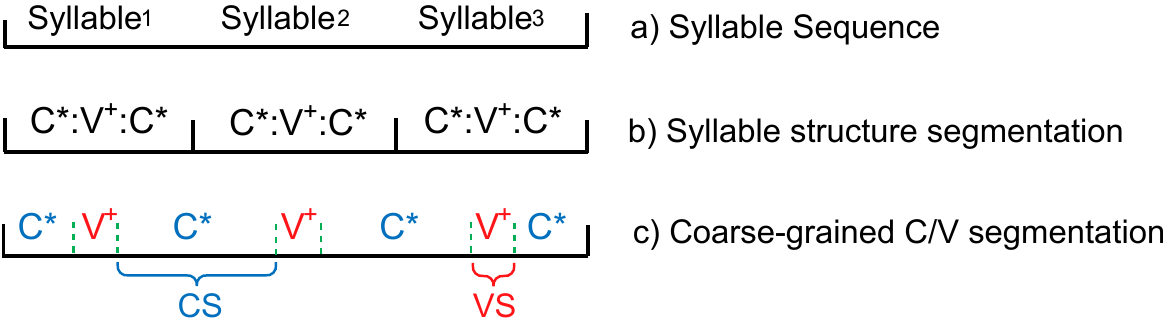}
 \end{figure}

Once the consonant/vowel segmentation is performed, our prosodic features are extracted at utterance level. Whereas, previous prosodic-based ADID researches  leveraging prosody  extracted prosodic features  at pseudo-syllable level ~\cite{rouas2006}~\cite{BiadsyH09prosody}.


 \section{Prosodic Statistic Analysis of Algerian Dialects }
 \label{sec:anovaProsodic}		

In this section, we first study whether Algerian Arabic dialects  can be discriminated using prosodic information.   Then, we discuss   what types of prosodic information can support Algerian Arabic dialect identification. 

\subsection{Speech Material}

It is important to mention that  there is  no standard corpus available  for Algerian Arabic  dialects such as for some other Arab countries~\cite{Alsulaiman13}. For this reason, we have  collected  our own corpus {\ALG} ~\cite{bougrine2016toward}.  This corpus provides a representation of phonetic, prosodic and orthographic varieties of Algerian Arabic dialects. Its current version  contains five Arabic Algerian dialects: Pre-Hil\={a}l\={\i}, Hil\={a}l\={\i}, Sulaymite,  Ma'qilian, and Urban Completely Bedouin dialects. Table \ref{table_database-analysis} gives more details on the sample chosen for the current analysis. The total number of utterances is $1892$,  each one is
about $6$s duration in average.

\begin{table}[!h]
	\centering
	\footnotesize
	\begin{tabular}{p{2.3cm}p{1.1cm}p{1cm}p{0.8cm}p{1.1cm}p{3.6cm}}
 		\toprule 	
 		 Dialect & \#Utterances & \#Speakers & \# Male & \# Female &  Department   \\ 
 		\midrule
 	    
 	    Pre-Hil\={a}l\={\i} (PreH) &143 & 03 & 01  &02 & Tlemcen \\ 
 	    \midrule 
 	    
 	    Urban C-B  (UCB) & 393 & 09 & 04&05   &  Algiers, Annaba,  M\'{e}d\'{e}a,  Mostaganem, Oran  \\
 	    \midrule

 	    Hil\={a}l\={\i} (Hil) &657 &14 & 05 & 09 & Adrar, Djelfa, Gharda\"{i}a,  Laghouat \\
 	    \midrule

 	    Sulaymite  (Sul)&469 &10 &05 & 05 & El-Oued\\
 	    \midrule
 	    
 	    Ma'qilian (Maq)&230&  05 & - &05 & Sidi Bel Abb\`{e}s \\ 

 	    \bottomrule   
 	     
 	  Total &1892&  41 & 15  &26 &  \\ 

 	   \bottomrule   
 	   
 	\end{tabular}
	\caption{Speech Material Details.}
	\label{table_database-analysis}
\end{table}

The speakers are chosen from adult population with $18$ to $50$ years old. The $41$ talkers are native from their dialect region, and both parents of each speaker were also native from the same dialect region.  The speeches gather both spontaneous and sub-spontaneous utterances.
\subsection{Prosodic Statistic Analysis}

In this statistical analysis, we have considered  the prosodic features presented in the previous section. Their extraction is done after the coarse-grained segmentation. More details on used tools are presented in experimental Section \ref{experiments}. Table~\ref{table_prosodicfeaturesall} presents  the means for each rhythm and intonation measure for   Pre-Hil\={a}l\={\i}, Urban C-B,    Hil\={a}l\={\i},  Sulaymite,   and Ma'qilian  dialects.

\begin{table}[!h]
	\centering
	\footnotesize
	\begin{tabular}{p{2cm}  p{1.4cm}  p{1.4cm}  p{1.1cm}  p{1.1cm}   p{1.1cm}     } 
 	  \\
 	  \toprule

 	  Feature & Pre-Hil\={a}l\={\i} & Urban C-B &   Hil\={a}l\={\i}&  Sulaymite  &Ma'qilian \\ \midrule
 	
 	\% V  & \textbf{39.1} & 43.47 & \textbf{40.4} &  \underline{\textbf{44.1}} & 43.7 \\ 
 
 	$\Delta$C  & \underline{\textbf{86.5}} & 57.3 & \underline{\textbf{97.1}} & \textbf{45.2} & 62.0  \\ 
 	 
 	$\Delta$V & 38.8 & 41.3& 36.7 & \textbf{34.2} &  \underline{\textbf{43.2}} \\ 
 	 
 	 VarcosC & \underline{\textbf{71.0}} & 62.5 & 63.2 & \textbf{57.9} & 64.4 \\ 
 	 
 	 VarcosV & 52.7 & \underline{\textbf{54}} & \textbf{50.1} & \textbf{49.6} & 52.9  \\ 
 	 
 	 rPVI-C & \underline{\textbf{88.7}} & 61.6 & \underline{\textbf{99.8}} & \textbf{49.9} & 65.5   \\ 
 	 
 	 nPVI-V  & \underline{\textbf{53.9}} &  \underline{\textbf{53.6}} & 51.9 & \textbf{49.5} & 52.6 \\ 
 	 
 	 Speech Rate  & 6.3 & 6.7& 6.8 & \underline{\textbf{7.3}} &\textbf{6.1}  \\ \midrule

 	 Pitch Range  & 12.8 & 10.6 & 8.9&\textbf{7.7}  & \underline{\textbf{16.6}}\\
 	 
 	   Pitch Top     & \underline{\textbf{330.3}} & 280.2 &267.6 &\textbf{260.6}  &328.6 \\   	 
 	
 Pitch	Bottom 	& 156.3 & 152.8 & 159.8 & \underline{\textbf{167.1}}  &\textbf{127.1}\\
 	
 Pitch	Mean   & \underline{\textbf{231.9}} & \textbf{207.2}  & 206.7 & 215.2 &209.1 \\
 	
 Pitch	TrajIntra & 8.2 & 8.1  & \underline{\textbf{9.9}}  &\underline{\textbf{10.1}} &\textbf{7.1}\\
 	
 Pitch	TrajInter & 9.9  &  10.5 &9.9 &\underline{\textbf{12.3}} &\textbf{9.6} \\
 	
 	  \bottomrule  
 	
 	   \end{tabular}
	\caption{Mean of Each  Prosodic Feature for Arabic Algerian Dialects.}
	\label{table_prosodicfeaturesall}
\end{table}

Cross-dialectal comparison shows that the proportion of vocalic intervals (\%V) represents less than $50$\% of the total duration of an utterance in all dialects, this result support the claim of  Hamdi et al. \cite{Hamdi04}, that all Arabic dialects have  less than $50\%$ for the vocalic intervals (\%V).

Pre-Hil\={a}l\={\i} and Hil\={a}l\={\i} dialects exhibit the smallest proportion of vocalic intervals (\%V), the greatest variability in consonant interval duration ($\Delta$C, rPVI-C), and  in vocalic intervals (nPVI-V). This fact shows that  they have the highest degree of vowel reduction and more complex syllable structure. The findings of the current study are consistent with those of Mar{\c{c}}ais \cite{marçais1960Algeria} who found that  Pre-Hil\={a}l\={\i}  dialect characterized by the  presence of vowel reduction.

 However,  Sulaymite dialect shows the greatest proportion of vocalic intervals (\%V), the smallest variability in consonant interval duration ($\Delta$C, rPVI-C), and  in vocalic intervals (nPVI-V).  Thus,  Sulaymite dialect is characterized by the absence  of  vowel reduction and simple syllable structure. These findings further support the note of Hamdi et al. \cite{Hamdi05} that Tunisian (Sulaymite) dialect has a simple syllable structure.

According to the speech rate,  we noticed  a faster overall articulation rate for  Sulaymite dialect, whereas a slower articulation rate for Ma'qilian dialect.

In order to confirm these observations, we have performed   statistical analysis one-way ANalysis Of VAriance  (ANOVAs). In this analysis, for each  feature  we consider only the dialects with the  lower   and higher  mean values respectively.  The summary of the significant differences between Algerian dialects is  reported  in Table \ref{table_signfeatures}.

\begin{table}[!h]
	\centering
	\footnotesize
	\begin{tabular}{p{2cm}  p{4cm}p{2.5cm} p{2cm} } \scriptsize
 	  \\
 	  \toprule 
 	 
   Features & Significant Dialects Differences & F & p \\   \midrule

     	\%V 		& Sulaymite $>$ Pre-Hil\={a}l\={\i}& F (1,610) = 60.44 & $p<.00001$  \\ 
     	 		& Sulaymite $>$  Hil\={a}l\={\i} &  F (1,1124) = 38.65 & $p<.00001$ \\ \midrule
     	
     	$\Delta$C	& Pre-Hil\={a}l\={\i} $>$  Sulaymite   &     F (1,610) = 95.11 & $ p<.00001$           \\ 
                	& Hil\={a}l\={\i} $>$  Sulaymite   &    F (1,1124) = 27.55 &  $ p<.00001$           \\ \midrule
     	
     	$\Delta$V  &  Ma'qilian $>$  Sulaymite       &    F (1,697) = 33.74 &  $ p<.00001$     \\  \midrule
     	
     	VarcoC     &  Pre-Hil\={a}l\={\i} $>$     Sulaymite   &    F (1,610) = 71.97 &  $  p<.00001$         \\\midrule
     	
     	VarcoV     &   Urban C-B $>$     Sulaymite           &      F (1,860) = 15.03 &  $  p<.001$     \\ 
                   &   Urban C-B $>$     Hil\={a}l\={\i}            &   F (1,1048) = 12.86&  $ p<.001$     \\ \midrule
     	
     	rPVI-C     &  Pre-Hil\={a}l\={\i} $>$ Sulaymite               &     F (1,610) = 102.9 &  $ p<.00001$    \\ 
     	
                  &  Hil\={a}l\={\i} $>$ Sulaymite               &  F (1,1124) = 25.97&  $ p<.00001$       \\ \midrule
     	
     	nPVI-V     &  Pre-Hil\={a}l\={\i}$>$ Sulaymite               &   F (1,610) = 8.71&  $  p<.01$      \\ 
     	          &  Urban C-B $>$ Sulaymite               &     F (1,862) = 15.38&  $  p<.00001$    \\ \midrule
     	
     	Speech Rate&  Sulaymite $>$  Ma'qilian  & F (1,697) = 132&  $  p<.00001$   \\ \midrule

        Pitch Range  &	 Ma'qilian $>$  Sulaymite  & F (1,697) = 674.3&  $  p<.00001$  \\   \midrule
       
        Pitch Top    &   Pre-Hil\={a}l\={\i}  $>$  Sulaymite   &  F (1,610) = 87.89&  $  p<.00001$\\  \midrule

     	Pitch Bottom &   Sulaymite   $>$  Ma'qilian & F (1,697) = 142.5&  $ p<.00001$  \\ \midrule
     
     	Pitch Mean   &   Pre-Hil\={a}l\={\i}   $>$ Urban C-B &  F (1,534) = 19.75&  $  p<.0001$  \\\midrule
     
     	TrajIntra    &   Sulaymite$>$  Ma'qilian &  F (1,697) = 90.44&  $  p<.00001$  \\ 
     	             &   Hil\={a}l\={\i}  $>$  Ma'qilian &  F (1,885) = 47.73&  $  p<.00001$  \\ \midrule
     	
     	TrajInter    &   Sulaymite $>$ Pre-Hil\={a}l\={\i}&  F (1,610) = 35.09&  $ p<.00001$  \\ 
     		         &   Sulaymite $>$  Hil\={a}l\={\i}&  F (1,1124) = 57.77&  $  p<.00001$  \\

 	       \bottomrule   
 	   \end{tabular}
	\caption{Summary of the Significant Differences between 	Algerian Dialects.}
	\label{table_signfeatures}
\end{table}

The main observation is that  the most p-value are smaller than .05 for each ANOVAs, which means that prosodic features can be used for binary classification.  
The second main  observation is  that   Sulaymite dialect has presented   significant differences with all the other dialects.    In addition,  Sulaymite dialect shows significant differences with Pre-Hil\={a}l\={\i} and/or Hil\={a}l\={\i} dialects for the  three  consonant variability measures:  unnormalized one ($\Delta$C)  and both consonantal intervals (VarcoC, rPVI-C), the proportion of vocalic intervals (\%V), and  both pitch measures (Top, TrajInter).

The vocalic interval duration ($\Delta$V), speech rate and three pitch measures (Range, Bottom, TrajIntra)  exhibit a significant difference between  Ma'qilian  and  Sulaymite dialect.

 The results of Ramus et al.\cite{Ramus99} study showed that the combination of \%V and $\Delta$C provides the best  correlate feature to separate traditional rhythm categorizations of languages (stress-timed, syllabic-timed, mora-timed). Stress-timed language (English and Dutch)  has full and reduced vowels. However, syllable-timed language (French,  Spanish)  does not have vowel reduction. In other words, the  percentage of vocalic sequences (\%V) of  stress-timed language is smaller than in syllable-timed language. Moreover, $\Delta$C is larger in stress-timed language and reflect  more complex syllable structure.

\begin{figure}[!h] 
	\centering
	\includegraphics[width=0.65\linewidth]{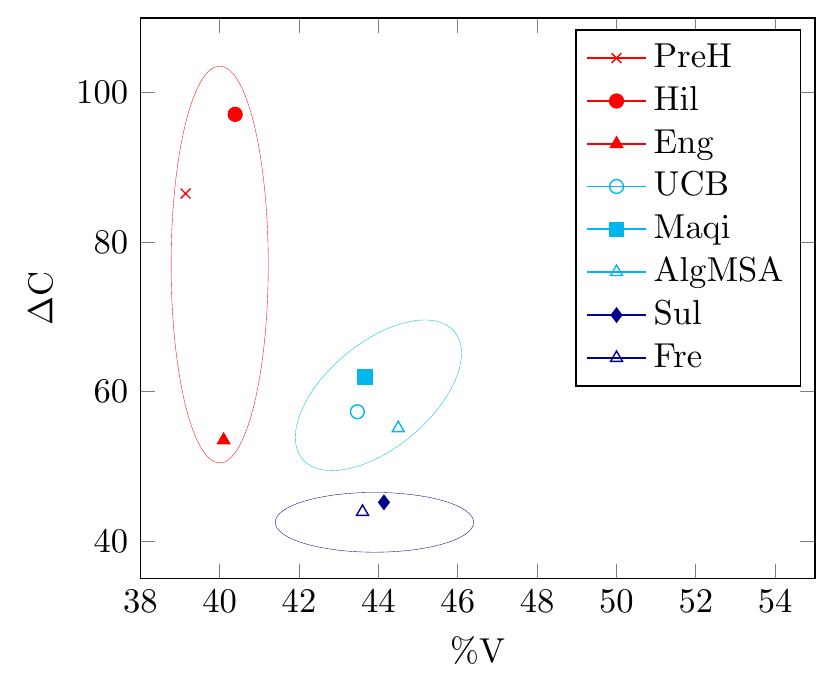}
	

	%
	%
	
	\caption{Distribution of Languages \& Dialects 
		along the \%V (x axis) and $\Delta$C Dimensions (y axis).} 
	\label{figurePerVDevC}
\end{figure}

 Figure \ref{figurePerVDevC} illustrates  the placement of the five Arabic Algerian  dialects with Algerian-MSA,  English, and French language  on the (\%V,  $\Delta$C) plane. The measures of Algerian MSA,  English, and French prosodic features are   taken from Droua et al. \cite{droua2010algerian}.

  From  the (\%V,  $\Delta$C) plane, we can divide   Algerian Arabic dialects into three groups. The first dialect group  has the highest $\Delta$C and the
  lowest \%V such as English were those traditionally classified
  as stress-timed. It include 	Pre-Hil\={a}l\={\i} and Hil\={a}l\={\i}  dialect.		The second group  represents the syllable-timed rhythm class, such as French, which include Sulaymite dialect. The third group includes  Ma'qilian and  Urban Completely Bedouin dialect that is near to Algerian-MSA language, which is    classified to  Mixed-timed rhythm class \cite{droua2010algerian}.

\begin{figure*}[t]
	

    	\resizebox{1\textwidth}{0.4\textwidth}{
    	\centering
    	\input{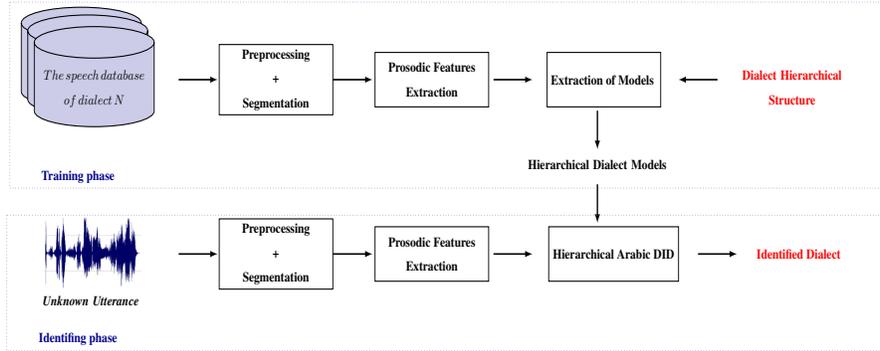} }
    
	\caption{Overview of our {\HD} system.}
	\label{view-approche}
\end{figure*}

\section{Hierarchical Dialect Identification  Approach}
\label{HADIDApp}

 Our proposed approach relies on prosodic features to identify Arabic  dialect in   intra-country context. In order to deal with the large dialectal varieties, our approach employs the dialect hierarchy structure to efficiently derive well training models.  Figure~\ref{view-approche}  sketches the global scenario of our HADID  approach.  As inputs, we provide a database of speech from different dialects. Within the preprocessing phase,  we remove noise and silence  from all the database speeches. Afterward, the coarse-grained consonant/vowel segmentation is applied and then  prosodic information are extracted. The latters processes are explained in   Section~\ref{prosoinfo}. For the extraction of  {\HD} models, we have used as an input our derived hierarchical structure for Algerian dialects which is presented   in Figure~\ref{class1}.

 For building {\HD} models, many hierarchical classification methods can be taken into consideration. The most used are  global or top-down  approaches~\cite{silla2011survey}. We opt to use the last one. It is called local classifiers approach, where the hierarchy is taken into account using local information.  More precisely, we built  a set of dialect models;  a Local Classifier per  Parent Node (LCPN)  from the top to down according to the  dialect hierarchical structure. We consider for each model only discriminative features. Hence, a different set of features and classifiers can be used separately for each node.   In other words, we choose among prosodic features those that have the best discriminative power for the embedded dialects by the node. Thus, each LCPN model dedicated to a node $X$  is trained using the   dataset that represents  dialects  for which $X$ is an ancestor.

  During the identification phase, an utterance   is determined by passing through the different DID systems within  the Hierarchical DID models. The target utterance is firstly classified to the most likely dialect group, proceeding level by level from the top until  the final dialect becomes identified.

   LCPN dialect models  are made using  the   state-of-the-art in language/dialect identification, Deep Learning technique \cite{LopezMoreno201646} \cite{Ferrer2016},  in this case  Deep Neural  Networks (DNNs).

DNNs  is an artificial neural network that has more than one layer of hidden units between its inputs and  outputs. The DNNs classifier  used in this work is a fully-connected feed-forward neural network with Rectified Linear Units (ReLU). Thus, an input $\mathtt x_{j}$ at level $\mathtt j$ is represented by Equation (\ref{eq:1}).  It is mapped to its corresponding activation $\mathtt y_{j}$  represented by Equation (\ref{eq:2}).

\begin{equation}
\mathtt x_{j}= b_{j} + \sum_{i} w_{ij}{\enspace}y_{i}
\label{eq:1}
\end{equation}
where $i$ is an index over the units of the layer below and $b_{j}$ is the bias of the unit $j$.

\begin{equation}
\mathtt y_{j} =  ReLU(x_{j})= \max(0, x_{j})
\label{eq:2}
\end{equation}

The output layer is then configured as a $\mathtt softmax$ function, where the hidden units map input $\mathtt y_{j}$ to a class probability $\mathtt p_{j}$ of the following form:

\begin{equation}
\mathtt{p_{j}=\dfrac{\exp (y_{j})} {\sum_{d}{\enspace}\exp(y_{d}) }} 
\end{equation}
where $\mathtt d$ is an index over all of the target dialect classes.

As a cost function for back-propagating gradients in the training stage, we use the cross-entropy function defined as:

\begin{equation}
\mathtt {C= -\sum_{j}t_{j}{\enspace}\log (p_{j})}
\end{equation} where $\mathtt t_{j}$ represents the target probability of the class $ \mathtt j$ for the current evaluated example, it takes a value of either $\mathtt 1$ (true class) or $\mathtt 0$ (false class).

\section{Experiments and Results}
\label{experiments}
In order to evaluate the performances of our  {\HD} approach, we have  performed various experiments. Firstly, we evaluate and analyze the performance of our {\HD}  system. Secondly, we measure the effect of using DNNs  for dialect modeling. In fact, we compare {\HD}  performance to  {\HD} system  that uses  SVM instead of DNNs modeling. Finally, we perform comparison of  {\HD}  with a baseline Flat classification approach, which is  built without considering  hierarchical dialect structure. Before presenting our experiments and results, we present the dataset, used tools, and evaluating metrics.


\subsection{Dataset and Tools}


Let us mention that  we   use the same corpus   as for the statistical analysis which is  presented in    Section \ref{sec:anovaProsodic}. It encompassed five dialects with 1892 utterances, each one is about 6s duration in average.  This
dataset is split into training (3/4) and testing  (1/4) sets.
 We note that we have performed speaker independent DID system. This means that the speaker set used for training  is disjoint  from speaker set used for testing which leads to real  dialect characterization capture.

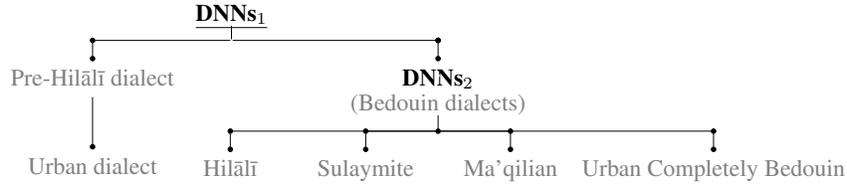
\begin{figure*}[t]
	\centering
	\scalebox{0.9} 
	{		\begin{tikzpicture}[-, scale=0.5]

		\tikzstyle{pet}=[circle,scale=0.1]
		\tikzstyle{inv}=[draw, circle, draw=black, fill=black, scale=0.4 ]
		\tikzstyle{inv1}=[draw, circle, draw=black, fill=black, scale=0.2]
		
		\tikzstyle{rec}=[draw, rectangle, draw=black, fill=black, scale=0.5 ]
		
		\tikzstyle{rec1}=[draw, semicircle, draw=black, fill=black, scale=0.35 ] 
		
		\tikzstyle{dia}=[draw, diamond, draw=black, fill=black, scale=0.3]
		\node[pet] (A) {};

		\node[pet] (A2) [ below=0.1cm of A,  scale=1, label=below:\underline{{\textbf{DNNs$_1$}}} ] {};
		\node[pet] (A21)[ below=0.4cm of A2] {};
		
		\node[pet] (B) [ below=0.2cm of A21 ] {};
		\node[inv1] (C) [ left=2cm of B ] {}; 
		\node[inv1] (D) [ right=3cm of B ] {}; 

		\node[inv1] (C1) [ below=0.2cm of C,scale=1, label=below:\textcolor{gray}{Pre-Hil\={a}l\={\i} dialect}] {};
		\node[pet] (C2) [ below=0.5cm of C1,  draw=white] {};
		\node[inv1] (C3) [ below=0.7cm of C2, scale=1, label=below:\textcolor{gray}{Urban dialect} ] {};

	    \node[inv1] (F) [ right=3cm of B ] {};

		\node[inv1] (F1) [ below=0.2cm of F,scale=1, label=below:\textbf{DNNs$_2$}] {};
		\node[pet] (F2) [ below=0.2cm of F1, draw=white ] {};
		
		\node[pet] (F22) [ below=0.1cm of F2,scale=1, label=below:\textcolor{gray}{(Bedouin dialects)}] {};
		\node[pet] (F222) [ below=0.4cm of F22, draw=white ] {};
			
		\node[inv1] (F3) [ below=0.2cm of F222] {};
		\node[inv1] (F4) [left=1cm of F3] {};
		\node[inv1] (F5) [ left=3cm of F3] {};
		\node[inv1] (F6) [ right=1cm of F3] {};
		\node[inv1] (F7) [ right=4cm of F3] {};
		
		\node[inv1] (FA) [ below=0.2cm of F5, scale=1, label=below:\textcolor{gray}{Hil\={a}l\={\i}} ] {};

		\node[inv1] (FB) [ below=0.2cm of F4, scale=1, label=below:\textcolor{gray}{Sulaymite}  ] {};

		\node[inv1] (FC) [ below=0.2cm of F6, scale=1, label=below:\textcolor{gray}{Ma'qilian} ] {};

			\node[inv1] (FD) [ below=0.2cm of F7, scale=1, label=below:\textcolor{gray}{{Urban Completely Bedouin}} ] {};

		\path

		(A21) edge node {} (B)
		
		(B) edge node {} (C)
		(B) edge [ below ] node {} (D) 
		
		(D) edge [ below ] node {} (F)

		(F) edge [ below ] node {} (F1)
		(F222) edge [ below ] node {} (F3)
		(F3) edge [ below ] node {} (F4)
		(F3) edge [ below ] node {} (F5)
		(F3) edge [ below ] node {} (F6)
		(F3) edge [ below ] node {} (F7)

		(F5) edge [ below ] node {} (FA)

		(F4) edge [ below ] node {} (FB)
		
		(F6) edge [ below ] node {} (FC)

			(F7) edge [ below ] node {} (FD)

		(C1) edge [ below ] node {} (C)

	    (C2) edge [ below ] node {} (C3)	
		
		;

		
		%
		\end{tikzpicture} 
	}
	\caption{Hierarchical Dialect Models.}
	\label{HADIDmodels}
\end{figure*}

In order to implement and evaluate our system, we have involved a set of Open Source softwares.
We have used Praat tool\footnote{Praat~v~$5.3.47$, Online: \url{http://www.fon.hum.uva.nl/praat}} to remove  noise. Our coarse-grained segmentation  is done automatically using also Praat  tool enhanced by   Prosogram\footnote{Prosogram~v~$2.9$, Online: \url{http://bach.arts.kuleuven.be/pmertens/prosogram/}} script. The latter  performs  syllable segmentation using intensity of band-pass filtered signal. The syllable' nuclei are delimited using spectral and amplitude changes. Please also note that this script  has reached  $80$\% according to   segmentation accuracy~\cite{salselas2011music}.

The  intonation metrics and speech rate  are extracted using  Prosogram script. While,   rhythm features  are calculated using Correlatore\footnote{Correlatore~v~$2.1$, Online: \url{http://www.lfsag.unito.it/correlatore/}} program, which is  mainly designed  for  rhythm analysis. 

Concerning the SVM generation of dialect models, we have  used    Weka tool\footnote{Weka~v~$3.7.13$, Online: \url{http://www.cs.waikato.ac.nz/ml/weka/}}, which is one of  the most commonly  tools in  Machine Learning.  The SVM kernel deployed is  a Radial Basis Function. We have got the best $C$ and $\gamma$ parameters of SVM. This is performed using GridSearch script, which is a meta-classifier~\cite{Braga2013}. In other side, for DNNs dialect modeling and test purposes, we have used  \emph{H2O}\footnote{H2O, Online: \url{http://www.h2o.ai/}} deep learning package scripted with \emph{R} language.

\subsection{{\HD} Implementation }

 According to the available hierarchical dialect structure, we have built two LCPN   based on DNNs classifier (DNNs$_{1}$, DNNs$_{2}$). Figure \ref{HADIDmodels} illustrates the  hierarchical dialect models.  DNNs$_{1}$ classifier is used to classify the first level dialect groups (Pre-Hil\={a}l\={\i}, Bedouin). It can be seen as a regional  dialect identification. Whereas, DNNs$_{2}$ classifier is used to classify  Bedouin dialects (Sulaymite, Ma'qilian, Urban Completely Bedouin, Hil\={a}l\={\i} dialect). 


 In order to choose  the appropriate prosodic feature set for each classifier, we have used the feature selection  method: ANOVA ranking \cite{wu2009feature}. This latter leads to the following:


\begin{figure}[t]
	
	\resizebox{1\textwidth}{!}{
		\centering
		\includegraphics{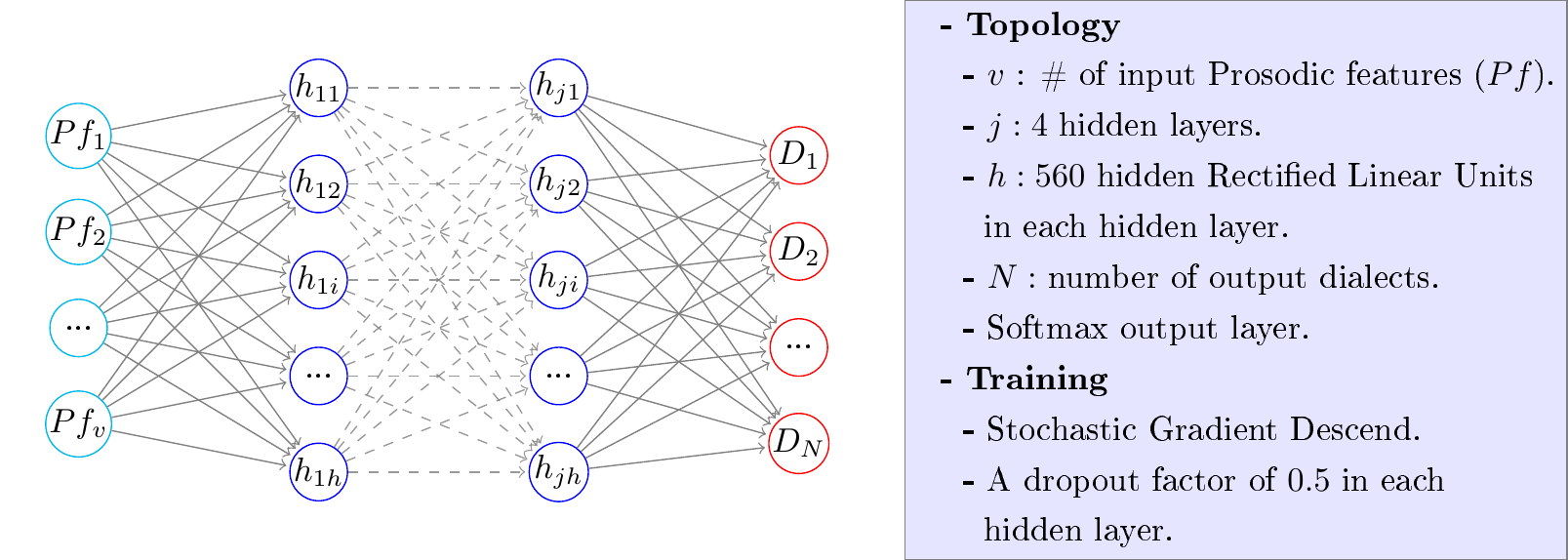}
	}
	\caption{DNNs Topology and Description. }
	\label{archDNN}
\end{figure}

  \begin{enumerate}
  	\item  For DNNs$_{1}$ classifier,  six features are selected.  These features are:  the duration proportion (\emph{\%V}),   the standard deviation of vocalic interval (\emph{$\Delta$V}), and four pitch values  (\emph{Range},  \emph{Mean},  \emph{TrajIntra},   \emph{TrajInter}). Thus, the input layer has $6$ linear units. Whereas,  the output layer has two units (Pre-Hil\={a}l\={\i},  Bedouin).

  	\item For DNNs$_{2}$ classifier, seven features are selected. These features are:   the duration proportion (\emph{\%V}),   the standard deviation of consonantal  interval (\emph{$\Delta$C}), and five pitch values (\emph{Range}, \emph{Top}, \emph{Bottom}, \emph{TrajIntra}, \emph{TrajInter}). Thus, the input layer has $7$ linear units. Whereas,  the output layer has  four units.
  \end{enumerate}

Now let us describe both topologies within DNNs$_{1}$ and DNNs$_{2}$. Figure \ref{archDNN} illustrates  the generic  DNNs topology and its related description. 
Each DNNs has four hidden layers with $560$ units. A dropout factor of $0.5$ is used for each hidden layer. These parameters are chosen empirically.




\subsection{Evaluation Metrics}
 Concerning the performance measure of {\HD} system, we use the extended version of the well known metric \emph{Precision}   but tailored to the hierarchical classification scenario namely  \emph{hierarchical   Precision} ($hP$) proposed by   Kiritchenko et al.~\cite{Kiritchenko:2006}. It is defined as follows: 
 
 {
 	$$ hP  = \frac{\sum_i|\hat{C_i}\cap\hat{T_i}|}{\sum_i|\hat{C}_i|}$$
 	} where $\hat{C_i}$  is the set consisting of the most specific  class(es) predicted for test example $i$ and all its ancestor classes, $\hat{Tˆ_{i}}$ is the set    consisting of the true most specific class(es) of test example $i$ and all its ancestor   classes. 

 	The performance of the whole  {\HD} system is measured using Hierarchical micro-precision, which is the average of  $hP$  on the entire test utterances. The performance of the baseline   Flat classification approach is measured using the standard
 	micro-Precision.  


\subsection{Results and Discussion }

For all  experiments, we use   speaker-independent DID  and the k-fold cross-validation technique ($k = 5$).  In order to ensure that our results are reliable.

\subsubsection{{\HD} System}
The discussion of the results begins with the study the performance of our {\HD} system. Table~\ref{HADIDkfold}  gives the Hierarchical micro-precision for {\HD} system 
on five folds.  

\begin{table}[h]
	\centering
	\small
	\scriptsize	
	\begin{tabular}{p{2cm}  p{2cm}  p{2cm} }
 	  \toprule 
             
	 	                     &  Level-1 (\%)	 &  Whole  System  (\%)             \\
 	    \midrule
 	    	Fold 1           & 88.3   	&  	64.0          \\  	    \midrule
 	    	
 	    	Fold 2  		 & 81.0	        &  65.0	         		\\  	    \midrule	
 	    	
  	    	Fold 3  		 & 86.0	    &  62.8	         		\\  	    \midrule	
  	    	
 	    	Fold 4  		 & 77.6	    &  57.6	         		\\  	    \midrule		   
 	    	
 	    	Fold 5  		 & 86.1	    &  65.0         		\\  	    \toprule	 
 	    	
 	    	Average  		 & 83.8	    &  62.8	         		\\  	        
 	       \bottomrule   
 	   \end{tabular}
 	   
	\caption{Our {\HD} System  Precision on Different Cross-validation Folds.}
	\label{HADIDkfold}
\end{table}

These results confirmed that prosody is suitable
to separate region dialects (Level-1). In fact, DNNs$_{1}$ classifier separates between Pre-Hil\={a}l\={\i} dialects and  Bedouin ones with 83.8\% of precision.
The average precision remains acceptable (62.8\%) for all the five dialects in spite of their closeness. We can also observe that the precision of the system is quite stable. In fact, the deviation between the precision for each fold  and the average precision doesn't exceed 5.2\%.

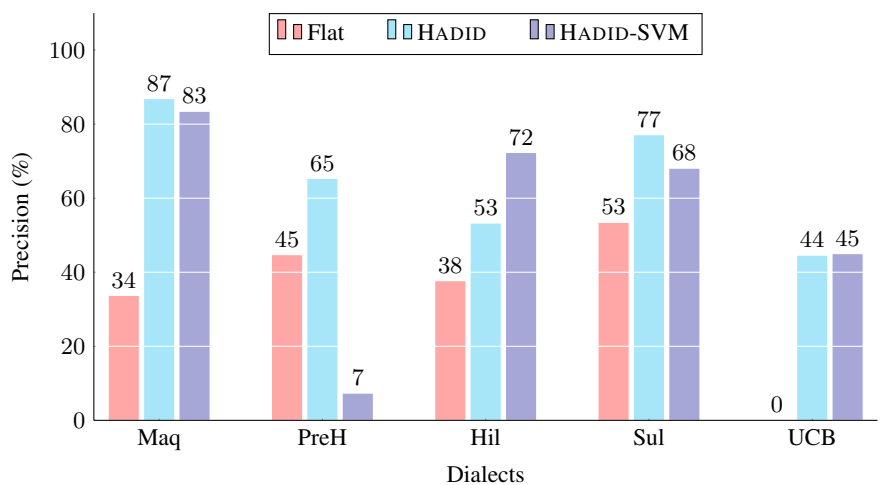
\begin{figure*}
	\centering
	\small
	\begin{tikzpicture}
	
	\begin{axis}[
	ybar, axis on top,
	height=7cm, width=12cm,
	bar width=0.4cm,
	ymajorgrids, tick align=inside,
	major grid style={draw=white},
	enlarge y limits={value=.1,upper},
	ymin=0, ymax=100,
	axis x line*=bottom,
	axis y line*=left,
	tickwidth=0.6pt,
	enlarge x limits=true,
	legend style={
		at={(0.5,1)},
		anchor=north,
		legend columns=-1,
		/tikz/every even column/.append style={column sep=0.5cm}
	},
	ylabel={Precision (\%)},
	xlabel={Dialects},
	symbolic x coords={Maq, PreH, Hil, Sul, UCB},
	xtick=data,
	nodes near coords={
		\pgfmathprintnumber[precision=0]{\pgfplotspointmeta}
	}
	]
	
	\addplot [draw=none,fill=red!35] coordinates {
		(Maq,33.58) (PreH,44.6) 
		(Hil,37.58) (Sul,53.32) (UCB,0)};
	
	\addplot [draw=none, fill=cyan(process)!35] coordinates {
		(Maq,86.74) (PreH,65.2) 
		(Hil,53.14) (Sul,77) (UCB,44.46)};
	
	\addplot [draw=none, fill=darkblue!35] coordinates {
		(Maq,83.32) (PreH,7.2) 
		(Hil,72.2) (Sul,67.94) (UCB,44.88)};
	
	\legend{Flat, \HD, {\HD}-SVM }
	\end{axis}

	\end{tikzpicture}
	\caption{Comparative Results of Flat Classification, {\HD},  and  {\HD}-SVM Systems by Dialect.}
	\label{comp3Approachs}
\end{figure*}

Figure~\ref{comp3Approachs} reports more details on   {\HD} system results. In fact,  it  reports average $hP$ by  dialect of the five folds.

 The best result for  {\HD} system is observed for Ma'qilian and Sulaymite dialects with 87\% and 72\%  precision respectively. This achievement  is predicted   by our statistical analysis.  The worst classification result is observed for the UCB dialect with 44\% precision. This result   can be explained by the fact that  UCB dialect has    Pre-Hil\={a}l\={\i} origins but deeply influenced   by Bedouin dialects. A deep examination of results shows that UCB utterances are mis classified in others Bedouin dialects. This fact suggests a refinement of the hierarchical dialect structure which needs dialectologists efforts.

\subsubsection{{\HD} vs. {\HD}-SVM System }

Turning now to the  comparison of {\HD} with {\HD}-SVM system. This latter designed as {\HD} system where dialect modeling uses conventional SVM classifier. Table \ref{DIDcompSVM} gives the comparative results in term of their $hP$ for 5-fold cross-validation. 

Concerning the first level, regional dialect identification,  {\HD}-SVM precision is of the same magnitude as {\HD} system. It is apparent from   that {\HD} (with DNNs) system is better than {\HD}-SVM system with an improvement of  7.1\% in term of precision. However, detailed results by dialects (see Figure~\ref{comp3Approachs})  proves that {\HD} is more suitable than {\HD}-SVM system. Indeed, the deviation between best and worst precisions by dialect is about 76\%    for {\HD}-SVM system.


\begin{table}[!h]
	\centering
	\small
	\scriptsize	
	\begin{tabular}{p{2.2cm}  p{2cm}  p{2cm} }
 	  \toprule 
             
	 	                 &  Level-1 (\%)	     & Whole  System (\%)	        \\
 	    \midrule

 	    	{\HD}-SVM		 & 83.0 	  	& 58.6  			\\  	    \midrule

 	    	\HD    & 83.8  		        & 62.8		 \\

 	       \bottomrule   
 	   \end{tabular}
	\caption{ {\HD} vs. {\HD}-SVM  Performance in Term of Precision.}
	\label{DIDcompSVM}
\end{table}

\subsubsection{{\HD} vs. Flat System}

The designed baseline Flat classification system  is built using the same preprocessing and segmentation phases  as for {\HD} approach. The speech is also characterized using the best prosodic features according to ANOVA ranking method.  However, one DNNs model is generated for all targeted dialects. 


 Table~\ref{Flat}  reports comparative results  of Flat classification and {\HD} system in term of precision.

 The average precision is about  38.4\% for Flat system. It is  clear that {\HD} outperforms this baseline Flat classification system by an improvement more than 63.5\%. This main finding  proves that for ADID, Hierarchical classification is more suitable than Flat classification. 
Furthermore, from Figure \ref{comp3Approachs}, we can observe  that UCB dialect is not at all classified  by Flat system. This is due to the same reason cited above.

\begin{table}[!h]
	\centering
	\scriptsize
	\begin{tabular}{p{1.5cm}  p{1cm}  p{1cm} p{1cm}  p{1cm} p{1cm}  p{1cm}}
 	  \toprule 
 		
	 	   System &  Fold 1 	 &  Fold 2    & Fold 3 	 &  Fold 4 & Fold 5     &   Average    \\
 	    \midrule
 	      Flat    &  43.7    &  	31.9	        & 35.7	 &  36.8	 & 43.9	 &  38.4	   \\  	    \midrule
 	    
 	    \HD    & 64.0  	&  	65.0	        & 62.8	 &  57.6 & 65.0 & 62.8	  \\


 	       \bottomrule   
 	   \end{tabular}
	\caption{Flat Classification vs. {\HD} System Precision on Different Cross-validation Folds. }
	\label{Flat}
\end{table}

%

\section{Conclusion}
\label{concl}

In this paper, first, we have shown by means of statistical analysis that prosody has a discriminative power for Algerian Arabic dialects. The prosodic features are  extracted at the utterance level after our coarse-grained consonant/vowel segmentation.

We have also  designed a prosody-based Hierarchical  Arabic Dialect IDentification ({\HD}) that identifies  Algerian Arabic dialects from a speech.  Within  {\HD} system, a top-down method is involved where a Local Classifier per  Parent Node (LCPN)  is built according to the given predefined hierarchical dialect  structure.  The LCPN dialects models are constructed using Deep Neural  Networks method.

{\HD} performances are evaluated on  Algerian Arabic dialects corpus with $1892$ utterances, 
each one is about $6$s duration in average. For region dialects identification, {\HD} reaches a precision of $83.6$\%  while for dialect identification it gives $62.8$\%. These results prove its suitability for  Arabic Dialect IDentification (ADID). Compared with Flat classification system,  {\HD}   gives an improvement of $63.5$\%  in term of precision.

To the best of our knowledge, in the context of  Arabic dialect identification,  this is the first investigation that applies the   hierarchical classification method where  Deep Learning technique  is deployed  to generate dialect models. Furthermore, for Algerian Arabic dialects, it is the first ADID system leveraging prosody.

Certainly  our  dataset covers only a part of Arabic Algerian dialects and the hierarchical  dialects structure  needs more refinement,  this framework can be considered as a kernel  for more complete systems. It can deal with  all  Arabic dialects identification, as long as, we have an efficient hierarchical dialect structure.


 Moreover, there are more than one possible ways to perform {\HD} implementation. We plan  to investigate the Global Hierarchical methods. In another ongoing work, we are  investigated a combined approach using prosody and acoustic/phonetic  speech information.

%


\section*{References}

\bibliography{input/mybibfile}

\end{document}